\documentclass[a4paper,twocolumn,10pt]{article}
\usepackage{eurosis}
\usepackage{url}
\usepackage{natbib}
\usepackage[T1]{fontenc}  
\usepackage{graphicx}
\usepackage{array}

\bibpunct[; ]{(}{)}{,}{a}{}{;}

\begin{document}

\title{ON THE PROBLEM OF ENTITY MATCHING AND ITS APPLICATION IN AUTOMATED
SETTLEMENT OF RECEIVABLES}

\author{Lukasz Czekaj, Tomasz Biegus, Robert Kitlowski\\
Szybkie Składki Sp.\ z o.o., Innowacyjna 1, Suwałki, Poland \\
email: \texttt{lukasz.czekaj@aidlab.com}\\
Stanisław Raczynski, 
Mateusz Olszewski, 
Jakub Dziedzic, 
Paweł Tomasik,\\
PICTEC,  al. Zwyciestwa 96/98, bud.\ 4, lok.\ B3.06, Gdynia 81-451, Poland\\
Ryszard Kozera, Alexander Prokopenya\\
Warsaw University of Life Sciences -- SGGW, Nowoursynowska str. 166, 02-787 Warsaw, Poland\\
Robert Olszewski\\
Warsaw University of Technology, Pl. Politechniki 1, 00-661, Warsaw, Poland}

\date{}

\maketitle

\thispagestyle{empty}


\begin{abstract}
This paper covers automated settlement of receivables in non-governmental organizations. 
We tackle the problem with entity matching techniques.  We consider setup, where base algorithm is used for preliminary ranking of matches, then we apply several novel methods to increase matching quality of base algorithm: score post processing, cascade model and chain model.
The methods presented here contribute to automated settlement of receivables, entity matching and multilabel classification in open-world scenario.
We evaluate our approach on real world operational data which come from company providing settlement of receivables as a service: proposed methods boost recall from $78\%$ (base model) to $>90\%$ at precision $99\%$.

\end{abstract}

\section{INTRODUCTION}
Automated settlement of receivables plays an important role in optimisation of the efficiency of finance processes. Incomplete remittance information leads to  many hours being spent trying to determine whose payment refer to specific liabilities~\cite{aiInvoiceSimplification}.
AI solutions may significantly improve payment settlement time
\cite{aiFlywire}, reduce cost and manual effort \cite{automationCachApplication}.

As the main motivation and playground for testing presented ideas, we used the case of automated settlement of receivables in non-government organizations (NGOs) in the form of sport clubs. However, this problem applies to any situation where many payments have to be assigned to many and unequal number of receivables, which are manually settled by human operators in enterprise resource planning (ERP) systems. The problem requires matching imperfect real-world data (originating from manual typing to bank transfer details) with proper members/debtors. The above task is further complicated by lack of data structure, misspellings, typos or missing data. 

In automated settlement of receivables, for a given moment of time, we have two datasets $A$ (bank transfers) and $B$ (organization members). 
Rows contains some attributes, e.g. rows in $A$ contains: sender, title, date, amount; rows in $B$ contains: member id, member name, member address, guardian name, duties. 
Our goal is to determine for each pair $(a,b)\in A \times B$ if the transfer $a$ should be assigned to the organization member $b$ in order to settle his/her liabilities. 
Such an assignment is called a match and we denote a set of matches as M and it's complement as U (the set of unmatched pairs)
The matching relationship is many-to-many since one transfer may be used to pay liabilities of several members (e.g. siblings) and there may be multiple payments for one member (e.g. membership fee and sports insurance). However we expect that matching relationship splits into many sparse and small components. We also assume, that the vast majority of transfer matches with at least one member.

Since the set of members may change between time moments (e.g. new members constantly join organization) we cannot directly predict member from transfer attributes: we expect different member set during model development and production. Therefore we tackle the problem in question using entity matching techniques \cite{recordLinkage, recordLinkageComparison}: we calculate similarity score $p(a,b)$ between transfer $a$ and member $b$, then we use $p(a,b)$ to rank members for given transfer and select members above given threshold as matching. More information about similarity learning may be found in \cite{similarityLearning, kernelLearning}.

In classical entity matching, calculation of similarity score splits into two steps.
First we transform a pair $(a,b)$ into features vector $(a,b)\mapsto F(a,b) = [F_1(a,b)), \ldots]$, where $F_i$ are some features derived from attributes of $a$ and $b$. 
Features vector express details of similarity:
a feature is set true if certain pair of attributes of $a$ and $b$ have similar values/match (in person matching problem feature may reflect if name, date of birth or address matches). In our case we have unstructured data (transfer title which may contain names, description, key words) therefore we have to use more sophisticated features. Functions $F_i$ incorporate also some transformations and standardisation, e.g. replace spelling variations / misspelling of commonly occurring words with standard spelling. In the second step we apply scoring function $C$ to obtain scores from similarity vector $C:F(a,b)\mapsto p(a,b)$. Scoring function expresses how informative is the given feature for matching assessment and how different features interact. Scoring function $C$ may be probabilistic model, machine learning model \cite{beyoundProbRecLink} or deep learning model \cite{deepmatcher,Li2021DeepEM, Zhao2019AutoEMEF} (with end-to-end models that do not have explicit feature vector calculation step). 


It was stated above that in automated settlement of receivables we have some extra information about matching structure (e.g., split into many sparse and small components). 
As we are strongly focused on very high precision of matching (precision $\geq 99\%$), we optimise recall for this level of precision. This is motivated by the observation that it requires much more work to find and fix wrong match than to manually provide missing ones.
The above aim makes difference from classical entity matching and here we try to exploit these features.
We propose and evaluate different methods of post-processing of similarity scores obtained from matching algorithm: local distance transformation and two stacked models: cascade and chain models. 

Proposed methods are inspired by multilabel classification \cite{multilabelClassificaion, multilabelClassificaionTutorial} (where it is possible to assign more than one label for the sample, i.e.\ the sample may belong to more than one class) and open-world classification \cite{OpenWorldLearning} (new classes may appear in test/production phase).

Described methods contribute to automated settlement of receivables, entity matching and multilabel classification in open-world scenario.

\section{DATASET}
The research was conducted with the aim of solving business problem of a company which provides settlement of receivables as a service. All the experiments were conducted on the dataset supplied by the company. Here we provide dataset characteristics.

Source dataset consists of data from $7$ clubs, it contains $~33000$ transactions, $~2300$ members and cover the period from $2018-02-05$ to $2020-09-10$. The source data set was cleaned. We removed duplicates (transactions were multiplied in source dataset when they were matched with multiple members or receivables), paypall, transactions which were not matched with any members by operator (e.g., direct donations) or there were modifications that made some data useless (e.g., ID has changed over time, missing members, broken relations id date base).
After such rectification we obtained about~$19000$ transactions. 

Transactions contain such fields as: sender, title, date of transaction, amount. Sender provides such information as name, surname and address. 
Title of transaction (string) contains various data (as substrings) since it is manually filled by the sender.
We can find there names $~71\%$, surnames $~70\%$, description $~68\%$. Title of transaction may contain some typos (e.g. there may be misspelling in name or some letters may be replaced with their Latin counterpart). Clubs ask payors to place in the title the unique ID of the member. ID has the form AA-BBBB where AA is the club's ID and BBBB is the numeric ID of member of the club AA. Payors are also asked to send only one transfer for member for a receivable. If these requirements were met, settlement of receivables would be easy. However as we will see this is not the case.
We observed that in general only about $~65\%$ of transfers contain IDs and there was $~4\%$ error rate in the ID format. There are $~5\%$ of transfers that are connected with more than $1$ member and $~13\%$ are connected with more than $1$ receivable. 

Due to the disproportion in representation of clubs in source data set (the biggest club has $~25$ times more transactions than the smallest) we have decided to balance dataset and sample at most $500$ transaction from each club. In that way we obtained dataset of $~3300$ transactions. 
The proportion of transactions with and without ID amounts to $47:53$

\section{METHODS}
In this section we provide description of our base matching algorithm, then we present our methods of increasing matching quality: score post-processing, cascade model and chain model.

Since clubs do not share accounts and receivables, we focus only on transactions and members from single club.
For every pair $(a,b)$ of bank transfer and club member we calculate features vector which consists of such features as: member ID in title, name in title/sender, member surname in title/sender, member guardian in title/sender, member/guardian address in sender. Because of declination in Polish language, we check for occurrence of grammatical forms of names and surnames. We also perform soft matching (e.g. matching with respect to Levenshtein distance) to cope with misspellings. 

We set the label (predicted value) for the pair $(a,b)$ to $1$ if it truly matches in the dataset. We use features vector and labels to train XGBoost model \cite{xgboost}. There is big disproportion between the number of positive and negative examples, but it affects only the bias/cutoff in XGBoost model. We do not need to re-balance or re-weight examples.

We evaluate matching quality in precision recall framework since matching is rare event (most of members do not match with given transaction). As it was explained, we focus on maximal recall for precision $95\%$ and $99\%$. We perform $5$-fold validation: $80\%$ of cases go to the train set and the remaining $20\%$ go to the test set.

{\it Our baseline} is raw XGBoost score. For a given transfer $a$ we accept as match all members $b$ from the same club as $a$ which obtain score above given threshold.

{\it The first method} which we use to increase quality of matching is score post processing. In this method we re-scale the scores in such a way that they sum to $1$ for given transaction $a$: $p(a,b) \mapsto p(a,b)/\sum_b p(a,b)$. We accept all the members above new (re-scaled) score. This corresponds with multilabel classification. The intuition behind this method is that we should not look on "absolute" matching score but rather on some "relative" score between the best matching member and other members. If there is no ID in title, for all members we have low scores, however there should be some gap between true matching and rest of pairs. In that case members with scores above the gap would be boosted and we accept the pair $(a,b)$ if this gap is big enough.

\begin{figure}[ht]
\includegraphics[width=8.5cm]{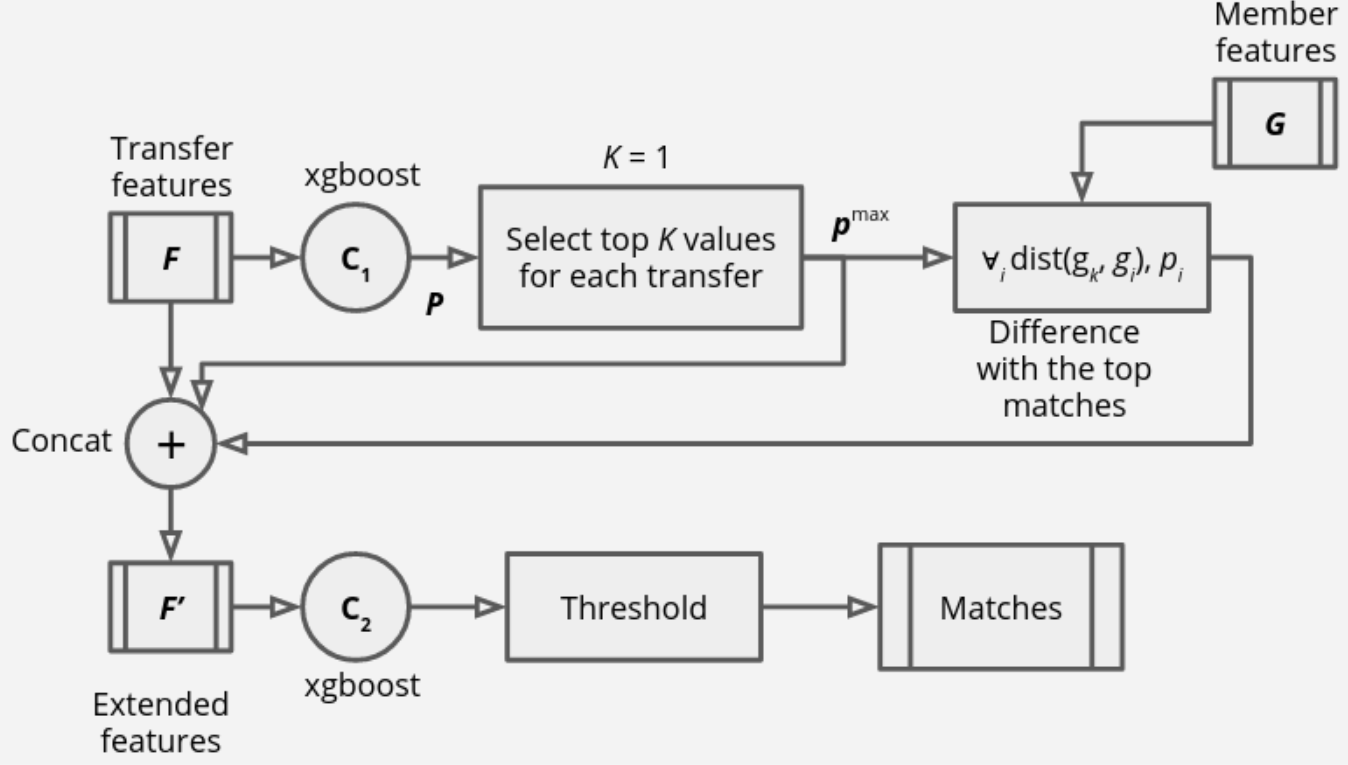}
\caption{Cascade classifier diagram, for detailed description see text. \label{cascade_classifier}}
\end{figure}

{\it The second method} is the cascade classifier (cf~\cite{chainClass}), which consists of two stages, as shown on FIG.~\ref{cascade_classifier}. Given features vector $F$ (calculated in the same way as in first method), the $C_1$ classifier determines the score $p$ of matching the transfer with a member. 
Then, for given transfer: the best member is selected (we denote its feature by $g_k$ and score by $p^{max}$), we compare every member of the club (we denote its features by $g_i$ and its score from $C_1$ by $p_i$) with best member; we obtain $dist(g_k, g_i) = g_k \oplus g_i$ vector of feature  which is binary comparison of $g_k$ and $g_i$. Finally we build extended features vector $F'$ which contains all features from $F$ with added following features: $p^{max}$, $dist(g_k, g_i)$, $p_i$, an integer signifying the position a member had in the ranking step of the previous classification, and the difference between the probability score for the member with higher and the lower position. The extended features vector is then reevaluated using classifier $C_2$ and obtained scores after threshold yields to match between transfer and member.

\begin{figure}[ht]
\includegraphics[width=8.5cm]{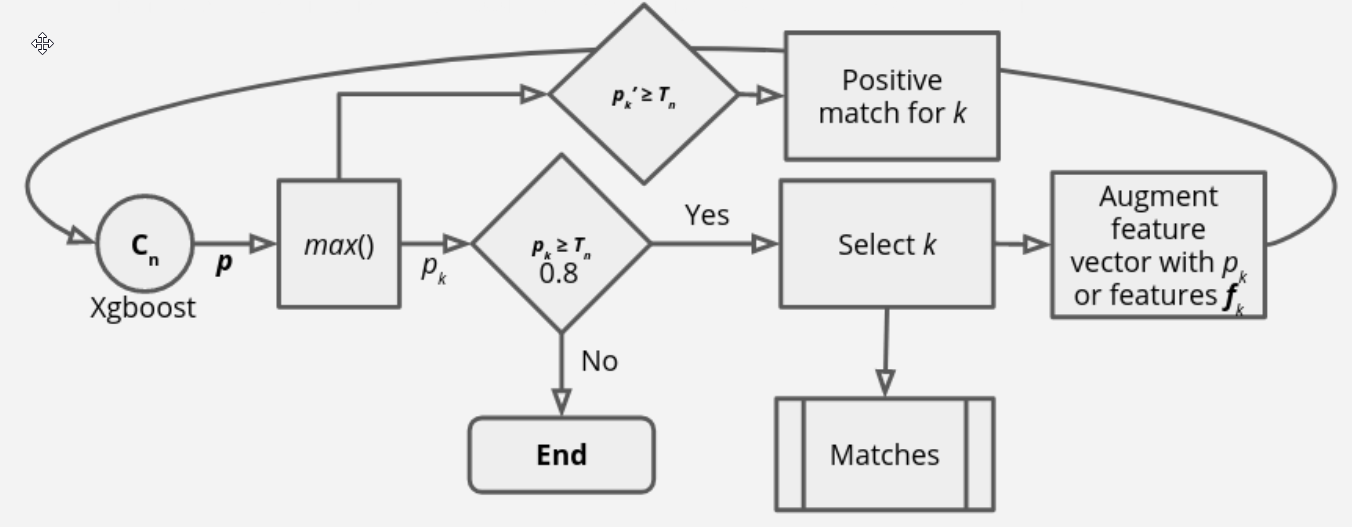}
\caption{Chain classifier diagram, for detailed description see text. \label{chain_classifier}}
\end{figure}

{\it The third method} is the chain classifier FIG.~\ref{chain_classifier}.
It works in iterative manner. It starts with base feature vector $F$ (the same as in previous methods). For given transfer, the best member is selected on the base of scores obtained from classifier $C_1$: if score is above threshold $T_1$, the members is accepted as matching to given transfer. If the best member matched, algorithm moves to the next iteration. In iteration $n$, best member from previous iteration is excluded from candidate members. The features of the remaining members are extended with the features of the best one from previous iteration. Algorithm uses classifier $C_n$ to obtain new scores. Algorithm finishes when no member gets score above $T_n$ or after $3$ iterations. In our experiments we use the same treshold $T_n = 0.8$.

\begin{figure}[ht]
\includegraphics[width=8.5cm]{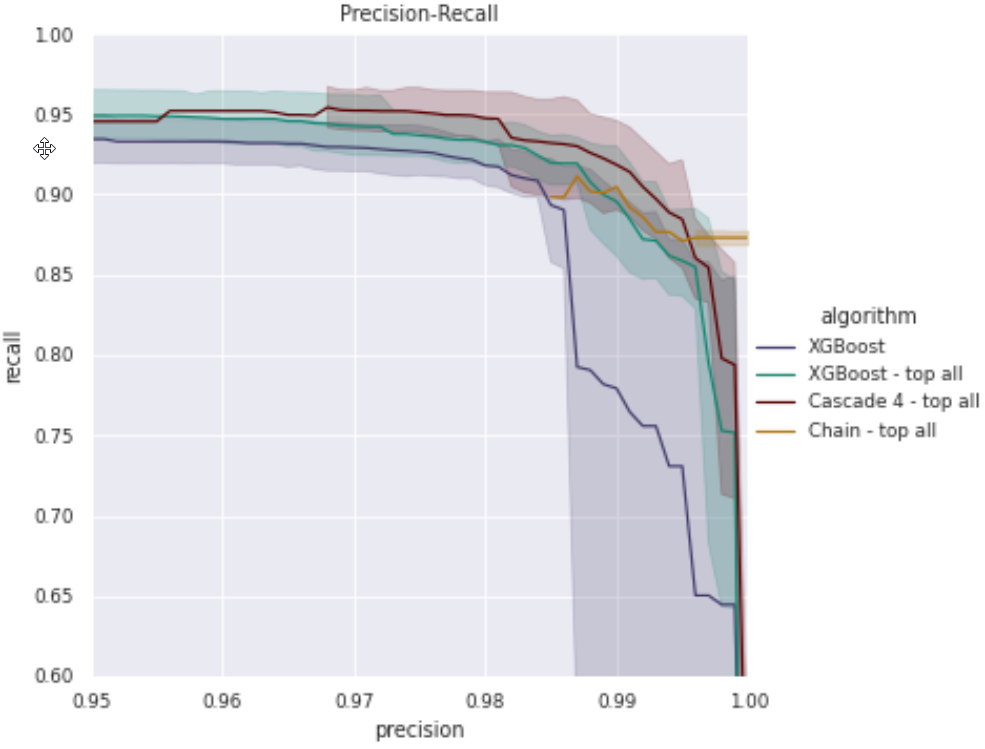}
\caption{Comparison of methods, all cases. \label{prec_rec_total}}
\end{figure}

Each of the models was evaluated using 5-fold cross-validation for the best model of each type. Best set of hyperparameters was found using grid search. Mean values of precision-recall curves alongside $95\%$ confidence intervals are presented on FIG.~\ref{prec_rec_total}. The plot shows performance of the best model of every descried method in the crucial range over $95\%$ precision.  
"XGboost" is a baseline method. "XGboost - top all" is composition of baseline method with score post processing. "Cascade 4 - top all" and "Chain - top all" are compositions of cascade/chain method with score post processing. We use this composition since they perform better than cascade/chain method alone.
As it was argued above, we focus on precision from interval $95\%$ - $100\%$ because it is easier to manually match transfer with member than look for false positive matchings.

\begin{figure}[ht]
\includegraphics[width=8.5cm]{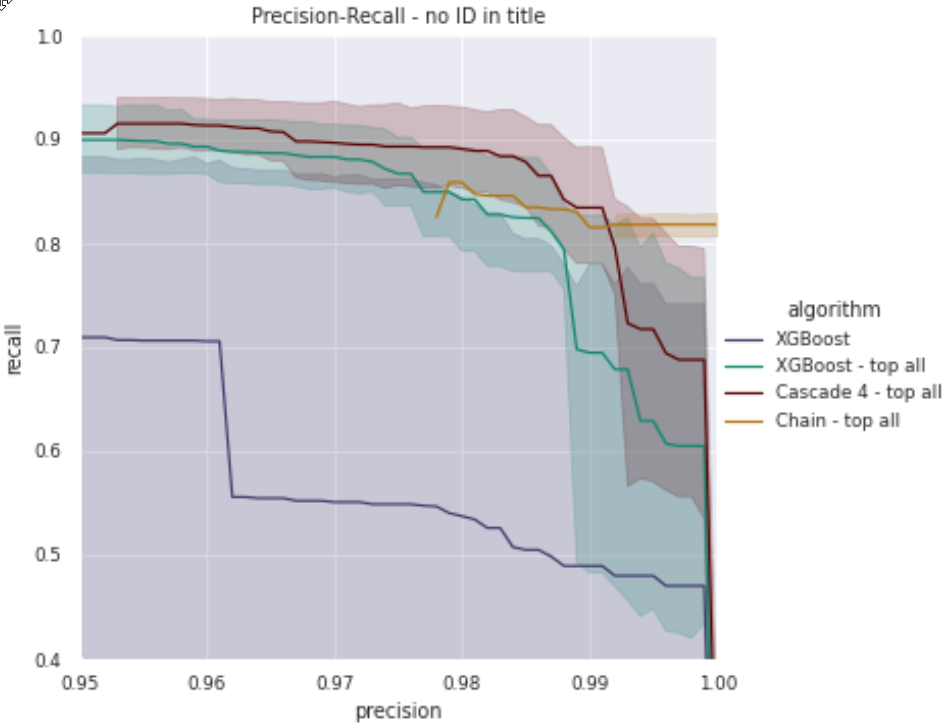}
\caption{Comparison of methods, no ID in title. \label{prec_rec_no_id}}
\end{figure}

To get better insight on the methods performance we provide separate figures for cases which do not contain ID in the title (FIG.~\ref{prec_rec_no_id}) and for the cases with ID in title (FIG.~\ref{prec_rec_no_id}). Detailed scores for two critical precision thresholds are summarized in TAB.~\ref{resutlst_tab}.


\begin{table*}[t]
\begin{center}
\begin{tabular}{ |m{3cm}|m{1.5cm}|m{1.5cm}|m{1.5cm}|m{1.5cm}|m{1.5cm}|m{1.5cm}| } 
 \hline
Method  & all ($95\%$) & all ($99\%$)& with ID ($95\%$) & with ID ($99\%$)& without ID ($95\%$) & without ID ($99\%$) \\ 
 \hline
 XGBoost & 0.934 & 0.779 & NA & 0.997 & 0.71 & 0.49\\ 
 XGBoost + top all & 0.949 & 0.9 & NA & 0.989 & 0.896 & 0.695\\ 
 Cascade 4 + top all & 0.945 & 0.918 & NA & 1.0 & 0.906 & 0.834\\ 
 Chain + top all & NA & 0.905 & NA & NA & NA & 0.816 \\ 
 \hline
\end{tabular}
\caption{Mean recall for best model in each category at 95\% and 99\% precision\label{resutlst_tab}}
\end{center}
\end{table*}

\begin{figure}[ht]
\includegraphics[width=8.5cm]{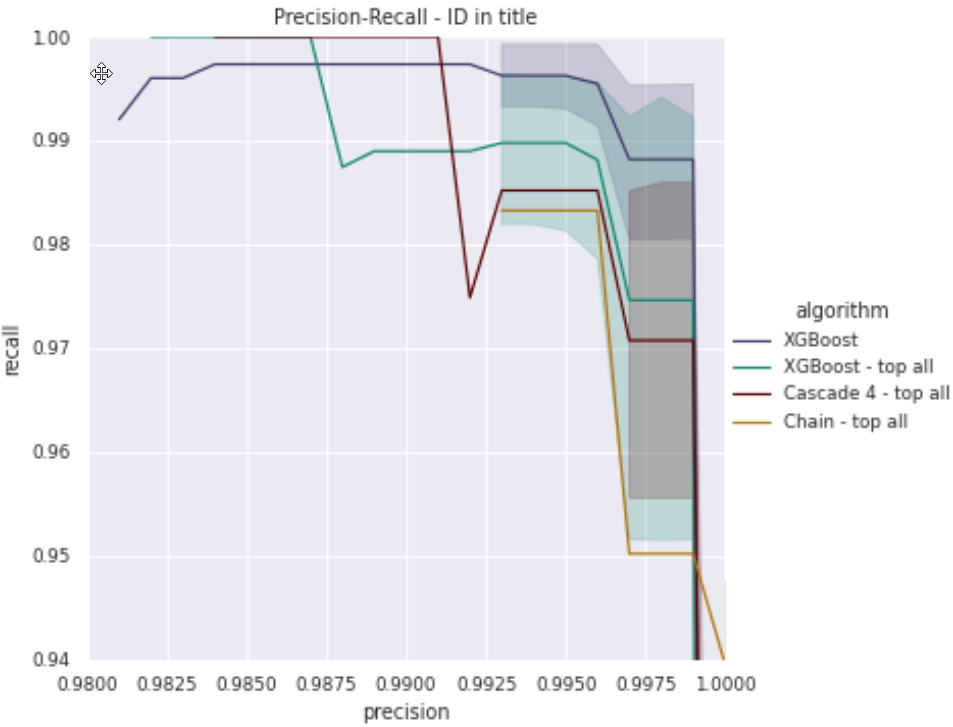}
\caption{Comparison of methods, ID in title.\label{prec_rec_with_id}}
\end{figure}


\section{RESULTS}
Baseline method is dominated by the proposed method in mixed case (transfers with and without ID in the title) and no ID in the title case. For the transfers that contain ID in title and extremely high precision (precision > $99\%$) we observe that baseline model provide the best results. Since detection of ID in the title is quite easy task, we can dispatch models if we need to operate in precision > $99\%$.

There is large variance in quality of baseline method for hard cases (no ID in title) according to cross-validation split. The reason is that in operating range (precision > $95\%$) miss-classification of few negative cases leads to significant change of precision and recall. "XGBoost - top all", "Cascade 4 - top all" and "Chain - top all" behave more stable than baseline methods. "Chain - top all" provide the most stable results.  

\section{DISCUSSION}
We have shown that entity matching framework together with the XGBoost classifier provides efficient machine learning model for automated settlement of receivables.
The model operates in a business motivated regime of high precision (>$95\%$).
We demonstrated several methods inspired by multi-label classification that increase quality of models in terms of precision-recall curve and cross-validation stability. 
Proposed methods are modular and may be combined with different base classifiers. We decided on XGBoost together with hand crafted feature because of relatively small dataset. Simpler models (logistic regression and Naive Bayes) was significantly worse.

Early results from the implementation of the presented methods in the company shows reduction of cost to time metric: we observed reduction of time required for settlement of receivables to $~25\%$ of the time originally spent on these task. The implementation may also reduce days sales outstanding metric, i.e. average number of days that receivables remain outstanding before they are collected or assigned to club member. Presented methods would be useful in cash flow management in NGOs.

Further experiments may show if additional threshold before score post processing helps: i.e. excluding from matching these transfers where best members has score below threshold. It may play role in case of missing members, e.g. first transfer came before the member was added to system.

Further research is required to determine in which cases the proposed methods help the most (score post-processing does not help if there is ID in title - very confident matching), when it is better to dispatch model and how proposed methods impact on models stability in context of cross validation.

Proposed methods may be also applied in more typical applications of entity matching: entity deduplication, record linkage, object identification.

\section{ACKNOWLEDGMENTS}
This research is a part of the project "Development of modern methods of data input and processing, automating the acquisition of funds ($\ldots$) based on artificial intelligence methods." no.\ POIR.01.01.01-00-0750/19

\bibliography{refs}

\end{document}